\pdfoutput=1

\documentclass[11pt]{article}

\usepackage[final]{acl}
\usepackage{times}
\usepackage{latexsym}

\usepackage[T1]{fontenc}
\usepackage[utf8]{inputenc}

\usepackage{amsmath}
\usepackage{amssymb}
\usepackage{microtype}
\usepackage{inconsolata}
\usepackage{enumitem}

\usepackage{graphicx}
\usepackage{caption}
\usepackage{booktabs}
\usepackage{multirow}

\usepackage{color}
\usepackage{xcolor}
\definecolor{light-gray}{gray}{0.95}
\usepackage{listings}
\lstdefinestyle{datastyle}{
  backgroundcolor=\color{light-gray}, % Use a light gray background, requires \usepackage{xcolor}
  basicstyle=\small\ttfamily,       % Use small, monospaced font
  breakatwhitespace=false,         % Don't break lines only at whitespace
  breaklines=true,                 % Break long lines automatically
  captionpos=b,                    % Puts caption at the bottom
  keepspaces=true,                 % Respects multiple spaces
  showspaces=false,                
  showstringspaces=false,
  showtabs=false,                  
  tabsize=2,
  frame=single,                    % Adds a frame around the box
  framerule=0.5pt,
  framesep=5pt,
  rulesepcolor=\color{gray}        % Color for the frame rules
}

\title{Exploring Gaps in the APS: \\
Direct Minimal Pair Analysis in LLM Syntactic Assessments}

\author{Timothy Pistotti \\
  University of Auckland \\
   \\\And
  Jason Brown \\
  University of Auckland \\
  \\\And
  Michael Witbrock\\
  University of Auckland
  }

\begin{document}
\maketitle
\begin{abstract}

Recent studies probing the Argument from the Poverty of the Stimulus (APS) have applied Large Language Models (LLMs) to test the learnability of complex syntax through surprisal-based metrics. However, divergent conclusions raise questions concerning the insights these metrics offer. While \citet{wilcox2024using} used direct minimal pair comparisons (the ``wh-effect'') to demonstrate that models successfully generalise knowledge of filler-gap dependencies, \citet{lan2024large} used a Difference-in-Differences (DiD) metric and found that models largely fail on parasitic gaps (PGs). This paper argues that the direct minimal pair approach offers greater diagnostic transparency. We demonstrate this by generating a full 8-permutation paradigm of refined PG stimuli and evaluating the GPT-2 model used in previous studies with a systematic Wilcox-style wh-effect analysis. Our results show that GPT-2 succeeds across all four tested conditions, indicating robust knowledge of filler-gap licensing principles even in complex PG environments. This finding, which contrasts with the more ambiguous results from DiD-style metrics, suggests that the choice of evaluation metric is critical for assessing an LLM's syntactic competence.

\end{abstract}

\section{Introduction}
\label{sec:introduction}

The evaluation of syntactic knowledge in Large Language Models (LLMs) has become a crucial area of research for understanding their capabilities and for empirically addressing foundational questions in linguistics, such as the Argument from the Poverty of the Stimulus (APS). Surprisal, the negative log probability of a word given its context, has emerged as a key psycholinguistic metric for these evaluations \cite{linzen-etal-2016-assessing, futrell-etal-2019-neural, wilcox2024using}.

Recent work has employed surprisal-based metrics to test LLM knowledge of complex dependencies, yet has adopted different evaluation paradigms. \citet{wilcox2024using} investigated various filler-gap dependencies by measuring a ``wh-effect,'' a direct surprisal comparison between minimal pairs that differ only in the presence of a \textit{wh-}filler versus a complementizer \textit{that}. Their findings generally indicated that LLMs successfully acquire knowledge of these structures.

In response, \citet{lan2024large} tested more complex phenomena---parasitic gaps (PGs) and across-the-board (ATB) movement. To do so, they introduced a Difference-in-Differences (DiD) metric, a statistical tool designed to measure an interaction effect across a $2 \times 2$ paradigm of stimuli. Their findings, showing poor LLM performance on PGs and ATB movement, were interpreted as support for the APS.

While both approaches have merit, this paper argues that they differ greatly in their diagnostic transparency. The direct minimal pair approach allows for clear, interpretable tests of specific linguistic hypotheses. We apply this more direct framework to the PG phenomenon and find that the model's knowledge is more robust than suggested by prior work, indicating that the choice of metric can significantly shape conclusions about model competence.

\section{Analysis of Evaluation Paradigms}
\label{sec:paradigms}

Though \citet{lan2024large} and \citet{wilcox2024using} both rely on surprisal-based evaluation of LLMs on syntactic phenomena, the specific comparisons made differ in their diagnostic power. Here, we detail the distinct approaches taken by each paper, summarised in Table~\ref{tab:metric_comparison}.

\begin{table*}[t]
\centering
\small
\begin{tabular*}{\textwidth}{@{\extracolsep{\fill}} l p{0.34\textwidth} p{0.51\textwidth} }
\toprule
\textbf{Paper} & \textbf{Prediction} & \textbf{Metric / Evaluation Method} \\
\midrule

\multirow{4}{*}{\shortstack{Wilcox et al.\\(2024)}} 
& 1. Gaps require an upstream filler. 
& 
\textbf{Wh-Effect (+gap):} The surprisal at post-gap material should be lower with a \textit{wh-}filler than with \textit{that}. 
\vspace{1mm}
\newline
\textit{Metric:} $S(w^{+}|C_{what}) - S(w^{+}|C_{that}) < 0$ \\
\cmidrule(lr){2-3}

& 2. Fillers require a downstream gap. 
& 
\textbf{Wh-Effect (-gap):} The surprisal at an overt NP filling a potential gap site should be higher with a \textit{wh-}filler than with \textit{that}. 
\vspace{1mm}
\newline
\textit{Metric:} $S(w^{-}|C_{what}) - S(w^{-}|C_{that}) > 0$ \\
\midrule

\multirow{4}{*}{\shortstack{Lan et al.\\(2024)}} 
& 1. An LLM should prefer the grammatical multi-gap PG structure over its ungrammatical counterpart where the main clause gap (G2) is filled. 
& 
\textbf{Direct Preference:} Compares the surprisal of the gapped vs. ungapped G2 continuation in a \texttt{+Filler} context. 
\vspace{1mm}
\newline
\textit{Metric:} $\Delta_{+\text{filler}} > 0$, where $\Delta = S(\text{ungapped}) - S(\text{gapped})$. \\
\cmidrule(lr){2-3}

& 2. The model's preference for a gapped G2 should be stronger when licensed by a \textit{wh-}filler than when it is absent.
& 
\textbf{Difference-in-Differences (DiD):} Compares the preference for a gap ($\Delta$) across \texttt{+Filler} and \texttt{-Filler} contexts. 

\vspace{1mm}

\textit{Metric:} $\Delta_{+\text{filler}} > \Delta_{-\text{filler}}$ \\
\bottomrule
\end{tabular*}
\caption{Comparison of core predictions and evaluation metrics. \citet{wilcox2024using} focus on direct minimal pairs where only the filler is manipulated. \citet{lan2024large} use a $2 \times 2$ paradigm to calculate an overall interaction effect (DiD).}
\label{tab:metric_comparison}
\end{table*}

The method used by \citet{wilcox2024using} relies on direct minimal pair comparisons where only a single variable is manipulated while the critical region remains identical. This approach offers high interpretability, as the resulting surprisal difference (the wh-effect) can be uniquely attributed to the model's reaction to the manipulated variable.

In contrast, the DiD metric employed by \citet{lan2024large} is necessitated by a paradigm where the critical words being compared are not identical. Here, direct comparison is confounded by the baseline lexical probabilities of the differing critical words. To illustrate, consider the representative example (item 2 from the \citet{lan2024large} project's dataset) shown in Table~\ref{tab:lan_confound_example}

\begin{table}[h!]
\centering
\small
\begin{tabular}{@{}llr@{}}
\toprule
\textbf{Condition} & \textbf{Critical Word} & \textbf{Surprisal (bits)} \\
\midrule
`+Filler, +Gap1, -Gap2' & ``you'' & 4.14 \\
`+Filler, +Gap1, +Gap2' & ``soon'' & 22.98 \\
\addlinespace[3pt]
`-Filler, -Gap1, -Gap2' & ``you'' & 5.77 \\
`-Filler, -Gap1, +Gap2' & ``soon'' & 23.34 \\
\bottomrule
\end{tabular}
\caption{Surprisal values for the critical word in each of the four conditions for ``I know\textbf{ who/that} Bob's talking to \textbf{(Jennifer)} is about to bother (\textbf{you}) \textbf{soon}.''}
\label{tab:lan_confound_example}
\end{table}

Calculating their direct preference metric, $\Delta_{+\text{filler}} = S(\text{you}) - S(\text{soon})$, yields a heavily skewed value of $4.14 - 22.98 = \mathbf{-18.84}$ bits. A large negative result like this, which may well result from the much lower frequency of the word ``soon'' than ``you'' in training data, makes it impossible to interpret the simple delta as a meaningful measure of syntactic preference. This is not an isolated case; out of the 8,064 items, we find an average baseline surprisal difference of approximately 11.5 bits between the adverbial (\texttt{gap}) and nominal (\texttt{-gap}) critical words across all conditions.

The DiD metric aims to resolve this issue by measuring the interaction effect, partially controlling for this baseline difference. However, this approach obscures the specific linguistic knowledge being tested. A large DiD effect shows that the model is sensitive to the filler's role, but does not, on its own, disentangle the distinct principles of PG licensing. This is further complicated by the fact that the `-Filler' conditions also manipulate the status of the G1 gap, preventing a clean baseline.

\section{Methods}
\label{sec:methods}

To achieve a more diagnostically precise evaluation of LLM knowledge of PGs, our approach centres on direct minimal pair comparisons. This requires a full set of stimuli to test the distinct syntactic constraints that constitute knowledge of the complex domain of parasitic gaps.

\subsection{Stimulus Dataset}
Using Gemini 2.5 as the generative model, we created a controlled dataset of 40 items (320 sentences total), containing all 8 permutations for each PG item given the variable conditions: $\pm$filler, $\pm$gap 1, and $\pm$ gap 2. The stimuli used unambiguous subject island structures (e.g., ``the story about \_'') and were manually vetted for pragmatic plausibility. From this set, 33 well-formed items (264 sentences) were used for analysis after excluding 7 for verb selection issues that rendered some conditions ungrammatical (see Appendix~\ref{sec:appendix_data} for a sample of the resulting data).

\subsection{Analytical Framework and Procedure}
Our framework applied the wh-effect metric ($S(\text{+Filler}) - S(\text{-Filler})$) across the four possible gap configurations present in our 8-permutation paradigm. This resulted in four direct minimal pair tests (P1--P4), outlined in Table~\ref{tab:wilcox_style_tests}. 

\begin{table*}[h]
\centering
\small
\begin{tabular*}{\textwidth}{@{\extracolsep{\fill}} l l p{0.4\textwidth} p{0.2\textwidth} }
\toprule
\textbf{Test} & \textbf{Gap Context} & \textbf{Minimal Pair Comparison} & \textbf{Expected Outcome} \\
\midrule

\textbf{P1} & \texttt{+G1, +G2} & `+F, +G1, +G2' vs. `*-F, +G1, +G2' & \\
\textit{Licensing} & \textit{(Full PG)} & \textit{Tests if the wh-filler licenses the full grammatical PG dependency compared to `that`.} & $S(\text{+F}) < S(\text{-F})$ \\
\addlinespace[3pt]

\textbf{P2} & \texttt{-G1, +G2} & `+F, -G1, +G2' vs. `*-F, -G1, +G2' & \\
\textit{Licensing} & \textit{(Simple Ext.)} & \textit{Tests if the wh-filler licenses a simple host gap (G2) when the parasitic gap (G1) is filled.} & $S(\text{+F}) < S(\text{-F})$ \\
\addlinespace[3pt]

\textbf{P3} & \texttt{+G1, -G2} & `*+F, +G1, -G2' vs. `*-F, +G1, -G2' & \\
\textit{Violation} & \textit{(PG, No Host)} & \textit{Tests the effect of a wh-filler when the host gap is filled, leaving an unlicensed PG.} & (Exploratory) \\
\addlinespace[3pt]

\textbf{P4} & \texttt{-G1, -G2} & `*+F, -G1, -G2' vs. `-F, -G1, -G2' & \\
\textit{Violation} & \textit{(No Gaps)} & \textit{Tests if a wh-filler creates surprisal when no gaps are available to be licensed.} & $S(\text{+F}) > S(\text{-F})$ \\

\bottomrule
\end{tabular*}
\caption{Proposed Wilcox-style minimal pair comparisons for parasitic gaps. Each test compares a `+Filler' sentence (`who') to a `-Filler' sentence (`that') while holding the gap configuration constant. The expected outcome refers to the surprisal at the identical critical region.}
\label{tab:wilcox_style_tests}
\end{table*}

The procedure was as follows: (1) We obtained BPE-level surprisals from GPT-2 for all 264 sentences. (2) Surprisals were aggregated for pre-defined critical regions by summing the surprisals of their constituent BPEs. A critical region was defined as the overt NP filling a gap (for `-gap' conditions) or the material immediately following the gap (for `+gap' conditions). (3) For each hypothesis (P1--P4), we calculated the per-item surprisal difference between the two sentences in the minimal pair. (4) one-sample t-tests were used to evaluate the significance of these mean differences.

\section{Results}
\label{sec:results}

We evaluated GPT-2 on our new dataset, using the 33 well-formed items that passed our grammaticality checks. This section first presents the results using the $\Delta$-based metrics before applying the more diagnostically expressive minimal pair framework.

\subsection{Applying Metrics from Lan et al. (2024) to the Dataset}
\label{ssec:lan_metrics_re-evaluation}

We calculated the direct preference ($\Delta_{+\text{filler}}$) and DiD using the four paradigm conditions corresponding to the $2 \times 2$ design. The accuracy scores are presented below, and visualised in Figure~\ref{fig:lan_accuracy_plot}.

\begin{itemize}[topsep=4pt, itemsep=0pt, leftmargin=2em]
    \item For the direct preference criterion ($\Delta_{+\text{filler}} > 0$), GPT-2 achieves an accuracy of only \textbf{51.5\%}, which is at chance level. The mean effect is positive but not statistically significant (Mean = 2.17 bits, t(32) = 1.49, $p = .072$).
    \item For the DiD criterion ($\Delta_{+\text{filler}} > \Delta_{-\text{filler}}$), GPT-2 achieves an accuracy of \textbf{87.9\%}. The mean DiD effect is large and highly significant (Mean = 5.17 bits, $t(32) = 7.11, p < .0001$).
\end{itemize}

While the highly significant DiD result might indicate that GPT-2 has acquired robust knowledge of PGs when tested on this dataset, the chance-level performance on the direct preference metric provides no real insight concerning the linguistic capabilities of the model.

\begin{figure}[h]
    \centering
    \includegraphics[width=0.9\linewidth]{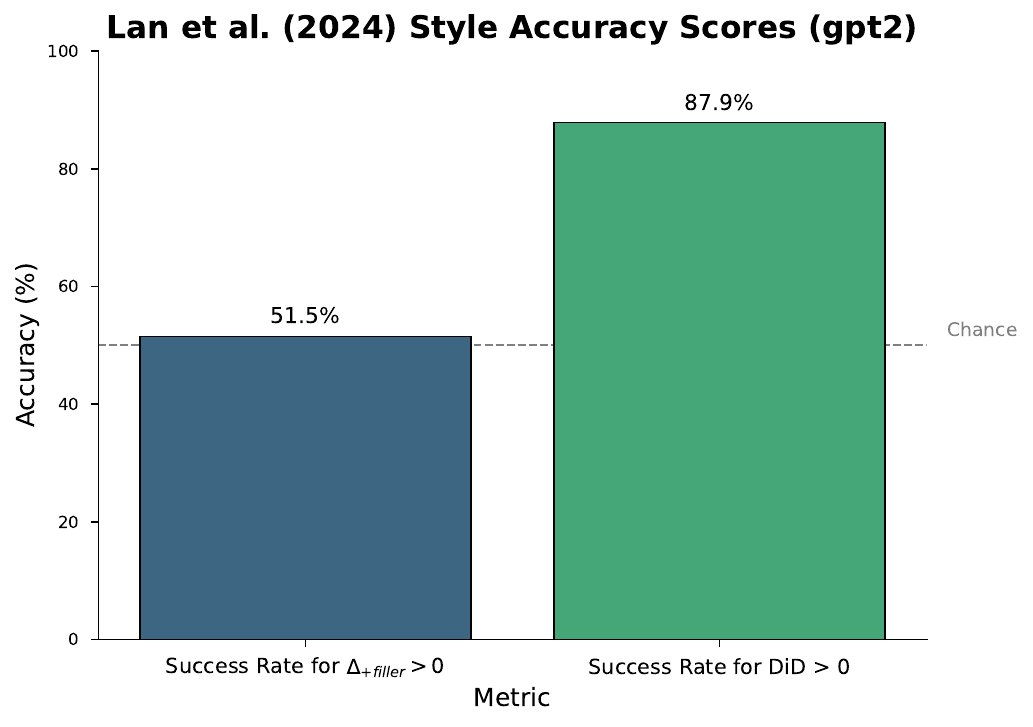}
    \caption{Results of Lan et al. metrics on our dataset}
    \label{fig:lan_accuracy_plot}
\end{figure}

\subsection{Fine-Grained Minimal Pair Analysis}

\begin{figure}[h]
    \centering
    \includegraphics[width=\columnwidth]{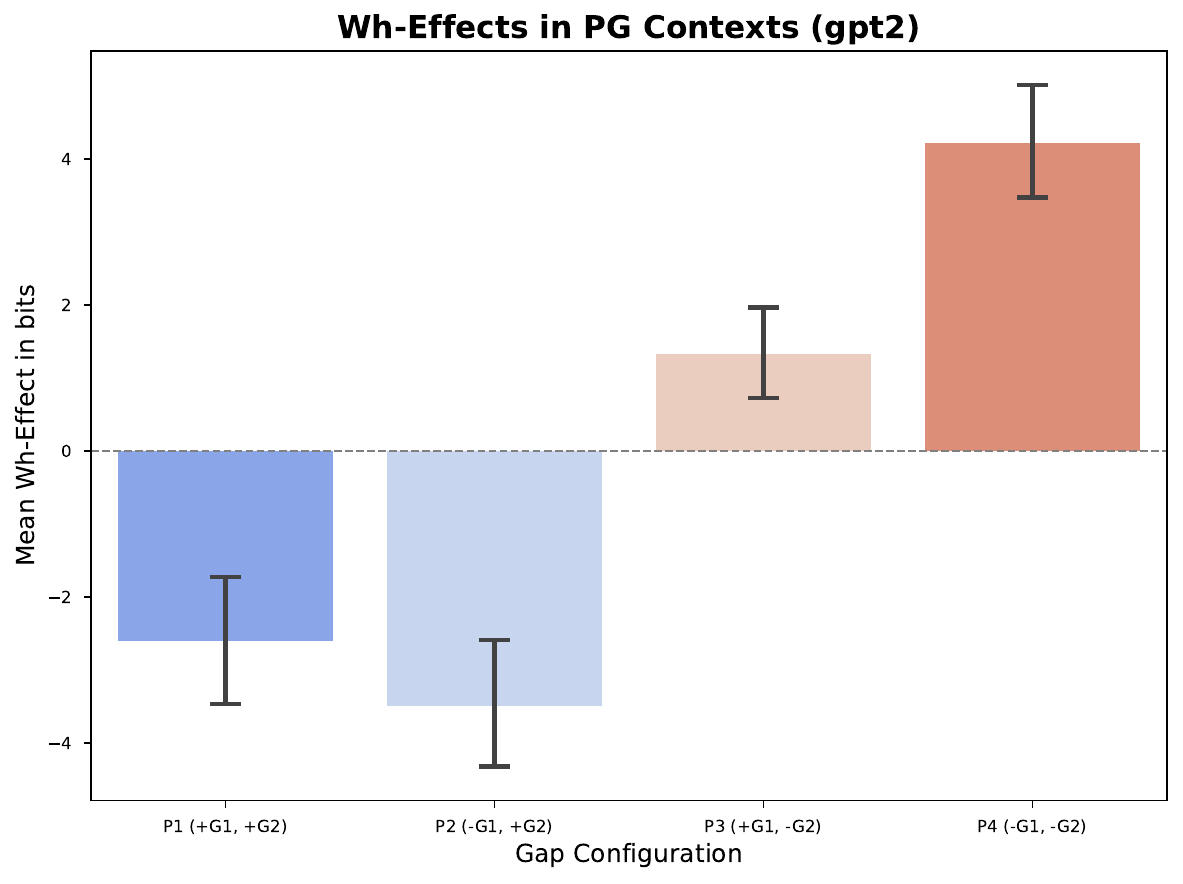}
    \caption{Mean wh-effects for the four gap configurations. Error bars represent 95\% confidence intervals. All effects are in the predicted direction and statistically significant.}
    \label{fig:wilcox_style_plot}
\end{figure}

We applied the Wilcox-style wh-effect analysis across the four gap configurations in our paradigm. The results, summarised in Table~\ref{tab:p1_p4_results} and visualised in Figure~\ref{fig:wilcox_style_plot}, reveal a consistent pattern of success.

\begin{table}[h!]
\centering
\small
\begin{tabular}{@{}lccc@{}}
\toprule
\textbf{Hypothesis} & \textbf{Mean (bits)} & \textbf{t-statistic} & \textbf{p-value} \\
\midrule
P1 (+G1, +G2) & -2.61 & -5.95 & \textbf{< .0001} \\
P2 (-G1, +G2) & -3.50 & -7.59 & \textbf{< .0001} \\
P3 (+G1, -G2) &  1.32 &  4.12 & \textbf{0.0002} \\
P4 (-G1, -G2) &  4.22 & 10.02 & \textbf{< .0001} \\
\bottomrule
\end{tabular}
\caption{Mean Wilcox-style wh-effects ($S(\text{+F}) - S(\text{-F})$) and statistics from one-sample t-tests (N=33 items). Significant results (p < .05) are in bold.}
\label{tab:p1_p4_results}
\end{table}

The results show a clear pattern of success. In the two grammatical licensing contexts, \textbf{P1} (full PG) and \textbf{P2} (simple extraction), the model correctly finds the sentences with a \textit{wh}-filler significantly less surprising than their ungrammatical counterparts with \textit{that}, as indicated by the large negative mean effects ($p < .0001$ for both).

Furthermore, in the two violation contexts, the model performs as expected. For \textbf{P4}, where there are no gaps to license, the model finds the sentence with a \textit{wh}-filler significantly more surprising than the grammatical baseline with \textit{that} ($p < .0001$). For the exploratory \textbf{P3} context, where the parasitic gap is unlicensed, the model also shows a significant positive wh-effect, robustly penalising the `+Filler' condition ($p = 0.0002$). These results indicate that GPT-2 has acquired a generalisable knowledge of filler-gap licensing that applies consistently across these complex structural variations.

\section{Discussion}
\label{sec:discussion}

Our fine-grained analysis, using direct minimal pair comparisons in the style of \citet{wilcox2024using}, reveals a consistent and surprisingly systematic knowledge of filler-gap dependencies in GPT-2, even within the complex syntactic environment of parasitic gaps (PGs). The model correctly distinguished grammatical from ungrammatical sentences across all four of our targeted licensing and violation contexts (P1--P4), with all effects being highly statistically significant.

This finding is particularly striking when contrasted with prior work. An unexpected outcome of our study emerged when we applied the $\Delta$-based metrics to our dataset. GPT-2's accuracy on the Difference-in-Differences (DiD) metric rose to \textbf{87.9\%} from the \textbf{68.8\%} reported by \citet{lan2024large} on their stimuli. Even more dramatically, the direct preference accuracy ($\Delta_{+\text{filler}}>0$) jumped from a reported \textbf{5.6\%} to \textbf{51.5\%} on our dataset.

We hypothesize that this marked improvement is not necessarily because the underlying linguistic challenge was simplified, but because the stimuli themselves are more representative of canonical PGs and free from specific confounds. This suggests that surprisal-based evaluations of complex syntax are highly sensitive to stimulus quality. The original conclusion that GPT-2 fails to learn PGs may have been at least partially influenced by unintended lexical ambiguities and structural complexities in the test data, rather than solely due to failure to acquire the syntactic generalisation itself.

This highlights the primary methodological takeaway: the choice of evaluation metric profoundly impacts the conclusions drawn about a model's capabilities. While a single interaction metric like the DiD can identify a general sensitivity to a licensor, it can obscure the details of what a model has learned. Our fine-grained P1-P4 analysis, by isolating specific linguistic principles, provides a more transparent and diagnostically powerful tool for building a more accurate picture.

\section{Conclusion}
\label{sec:conclusion}

This paper contrasted two prominent methods for evaluating LLM syntactic knowledge and argued for the superior diagnostic clarity of a fine-grained analysis based on direct minimal pair comparisons. Our results, using a new controlled dataset, indicate that GPT-2's knowledge of the principles governing parasitic gaps is more robust than previously shown. This suggests that conclusions about model capabilities are highly sensitive to both stimulus quality and the chosen evaluation metric.

We advocate that future research adopt more direct and interpretable tests. A logical next step is to apply this framework to the other models tested by \citet{lan2024large}, which performed even more poorly on the original dataset, to see whether performance there is similarly sensitive to stimulus design, or whether fine-grained analysis provides insights into what aspect of the PG licensing the model has failed to acquire. This approach promises a more rigorous foundation for claims about model capabilities and their implications for debates concerning the Argument from the Poverty of the Stimulus.

\bibliography{custom}

\begin{thebibliography}{4}
\providecommand{\natexlab}[1]{#1}

\bibitem[{Futrell et~al.(2019)Futrell, Wilcox, Morita, Qian, Ballesteros, and Levy}]{futrell-etal-2019-neural}
Richard Futrell, Ethan Wilcox, Takashi Morita, Peng Qian, Miguel Ballesteros, and Roger Levy. 2019.
\newblock \href {https://doi.org/10.18653/v1/N19-1004} {Neural language models as psycholinguistic subjects: Representations of syntactic state}.
\newblock In \emph{Proceedings of the 2019 Conference of the North {A}merican Chapter of the Association for Computational Linguistics: Human Language Technologies, Volume 1 (Long and Short Papers)}, pages 32--42, Minneapolis, Minnesota. Association for Computational Linguistics.

\bibitem[{Lan et~al.(2024)Lan, Chemla, and Katzir}]{lan2024large}
Nur Lan, Emmanuel Chemla, and Roni Katzir. 2024.
\newblock Large language models and the argument from the poverty of the stimulus.
\newblock \emph{Linguistic Inquiry}, pages 1--28.

\bibitem[{Linzen et~al.(2016)Linzen, Dupoux, and Goldberg}]{linzen-etal-2016-assessing}
Tal Linzen, Emmanuel Dupoux, and Yoav Goldberg. 2016.
\newblock \href {https://doi.org/10.1162/tacl_a_00115} {Assessing the ability of {LSTM}s to learn syntax-sensitive dependencies}.
\newblock \emph{Transactions of the Association for Computational Linguistics}, 4:521--535.

\bibitem[{Wilcox et~al.(2024)Wilcox, Futrell, and Levy}]{wilcox2024using}
Ethan~Gotlieb Wilcox, Richard Futrell, and Roger Levy. 2024.
\newblock Using computational models to test syntactic learnability.
\newblock \emph{Linguistic Inquiry}, 55(4):805--848.

\end{thebibliography}

\newpage
\onecolumn
\appendix

\section{Sample of Generated Stimuli}
\label{sec:appendix_data}

Below is a sample of our generated dataset presented in comma-separated value format. Note that Item 2, which uses the anti-rogative ``believed,'' is included here as an example of one of the 7 item sets excluded from our final analysis. This item was excluded because the main verb does not license a \textit{wh}-complement, rendering the `+filler' conditions ungrammatical and thus unsuitable for the intended minimal pair comparisons.

\begin{lstlisting}[style=datastyle, caption={Sample of Gemini 2.5 Generated Data}, label=lst:prompt]
sentence_type,item_id,condition,full_sentence
subject_pg_full,1,plusF_plusG1_plusG2,The investigators know who the story about is likely to damage severely.
subject_pg_full,1,plusF_plusG1_minusG2,The investigators know who the story about is likely to damage the campaign severely.
subject_pg_full,1,plusF_minusG1_plusG2,The investigators know who the story about the politician is likely to damage severely.
subject_pg_full,1,plusF_minusG1_minusG2,The investigators know who the story about the politician is likely to damage the campaign severely.
subject_pg_full,1,minusF_plusG1_plusG2,The investigators know that the story about is likely to damage severely.
subject_pg_full,1,minusF_plusG1_minusG2,The investigators know that the story about is likely to damage the campaign severely.
subject_pg_full,1,minusF_minusG1_plusG2,The investigators know that the story about the politician is likely to damage severely.
subject_pg_full,1,minusF_minusG1_minusG2,The investigators know that the story about the politician is likely to damage the campaign severely.
subject_pg_full,2,plusF_plusG1_plusG2,The audience believed who the picture of might have flattered greatly.
subject_pg_full,2,plusF_plusG1_minusG2,The audience believed who the picture of might have flattered the director greatly.
subject_pg_full,2,plusF_minusG1_plusG2,The audience believed who the picture of the actor might have flattered greatly.
subject_pg_full,2,plusF_minusG1_minusG2,The audience believed who the picture of the actor might have flattered the director greatly.
subject_pg_full,2,minusF_plusG1_plusG2,The audience believed that the picture of might have flattered greatly.
subject_pg_full,2,minusF_plusG1_minusG2,The audience believed that the picture of might have flattered the director greatly.
subject_pg_full,2,minusF_minusG1_plusG2,The audience believed that the picture of the actor might have flattered greatly.
subject_pg_full,2,minusF_minusG1_minusG2,The audience believed that the picture of the actor might have flattered the director greatly.
subject_pg_full,3,plusF_plusG1_plusG2,The board understood who the critique of would probably anger immensely.
subject_pg_full,3,plusF_plusG1_minusG2,The board understood who the critique of would probably anger the CEO immensely.
subject_pg_full,3,plusF_minusG1_plusG2,The board understood who the critique of the new project would probably anger immensely.
subject_pg_full,3,plusF_minusG1_minusG2,The board understood who the critique of the new project would probably anger the CEO immensely.
subject_pg_full,3,minusF_plusG1_plusG2,The board understood that the critique of would probably anger immensely.
subject_pg_full,3,minusF_plusG1_minusG2,The board understood that the critique of would probably anger the CEO immensely.
subject_pg_full,3,minusF_minusG1_plusG2,The board understood that the critique of the new project would probably anger immensely.
subject_pg_full,3,minusF_minusG1_minusG2,The board understood that the critique of the new project would probably anger the CEO immensely.


\end{lstlisting}

\end{document}